\documentclass[11pt,a4paper]{article}
\usepackage[hyperref]{emnlp2020}
\usepackage{times}
\usepackage{latexsym}

\usepackage{microtype}

\usepackage{graphicx}
\usepackage{booktabs} 
\usepackage{adjustbox} 
\usepackage{graphicx}  
\usepackage{multirow}
\usepackage{amsmath}

\aclfinalcopy 

\title{Learning by Semantic Similarity Makes Abstractive Summarization Better}

\author{Wonjin Yoon$^{1}$ \quad Yoon Sun Yeo$^{1}$ \quad Minbyul Jeong$^{1}$ \quad Bong-Jun Yi$^{2}$ \quad Jaewoo Kang$^{1}$ \\
Korea University$^{1}$, NAVER$^{2}$ \\
\texttt{\{wjyoon, ysy, minbyuljeong, kangj\}@korea.ac.kr} \\
\texttt{bongjun.yi@navercorp.com}
}

\date{}

\begin{document}
\maketitle

\begin{abstract}

By harnessing pre-trained language models, summarization models had rapid progress recently. However, the models are mainly assessed by automatic evaluation metrics such as ROUGE. Although ROUGE is known for having a positive correlation with human evaluation scores, it has been criticized for its vulnerability and the gap between actual qualities. In this paper, we compare the generated summaries from recent LM, BART, and the reference summaries from a benchmark dataset, CNN/DM, using a crowd-sourced human evaluation metric. Interestingly, model-generated summaries receive higher scores relative to reference summaries. 
Stemming from our experimental results, we first argue the intrinsic characteristics of the CNN/DM dataset, the progress of pre-trained language models, and their ability to generalize on the training data. 
Finally, we share our insights into the model-generated summaries and presents our thought on learning methods for abstractive summarization.
\end{abstract}

\section{Introduction}

Text summarization is a process of automatically generating a compact summary from a document while minimizing the loss of important information. 
There are two dominant methods for text summarization- namely \textit{Extractive} and \textit{Abstractive}.
Extractive summarization is a method of creating summaries by extracting important parts from the document \citep{zhang-etal-2018-neural, narayan2018ranking, liu-lapata-2019-text}, whereas abstractive summarization is more like generating sentences using salient information from the document \citep{see2017get, paulus2018deep, lewis2019bart}.

For abstractive summarization, reinforcement learning (RL) and supervised learning model are widely used. 
Reinforcement Learning (RL) based models, using ROUGE metric \cite{paulus2018deep} or neural network \cite{boehm_emnlp2019_summary_reward} as reward, showed remarkable performance. 
However, the optimization is slow and requires considerable computational effort to converge \citep{chen2018adversarial, gehrmann-etal-2018-bottom}. 
The supervised learning approach is straight-forward and requires relatively less time to train \cite{you-etal-2019-improving, gehrmann-etal-2018-bottom}.

Recently, large-scale pre-trained language models (LM), such as ELMo~\cite{Peters:2018}, BERT~\cite{devlin2018bert}, GPT-2~\cite{radford2019language} and BART~\cite{lewis2019bart}, demonstrated benefits of contextualized language representations through diverse natural language processing (NLP) tasks, including text summarization \citep{liu-lapata-2019-text, zhang-etal-2019-pretraining}. 
BART, which is composed of bidirectional transformers (encoder) and auto-regressive transformers (decoder), is designed as a sequence-to-sequence model and has shown advanced performance for CNN/DailyMail dataset.

However, most papers evaluated their model with automatic evaluation metrics such as ROUGE or METEOR.
Although automatic evaluation metrics are known to be have high correlation with human judgement, they have limitations and may mislead the actual quality of a generated text \cite{schluter2017limits, maynez2020faithfulness, gehrmann2021gem}.
Despite the fact that existing researches suggest these drawbacks of current evaluation strategy, papers with comparative analysis on the model-generated summaries and the reference summaries are, to the best of our knowledge, limited.

In this paper, we first investigate and compare the quality of the reference summaries and model generated summaries using crowd-sourcing based manual evaluation, or \textit{human evaluation}.  
The human evaluation result indicates that there is a statistically significant difference between reference summaries and the model generated summaries; interestingly, the latter was better.
Normally, quality of reference dataset become a upper-bound for a model trained on a given dataset; model shows sub-optimal performance than its own reference. However, our crowd-source workers preferred model-generated summaries over the reference summaries.

In the rest of our paper, we first address this observation, by discussing the intrinsic characteristics of CNN/DM - how the dataset was built. We also discuss the advance of pre-trained language model structure, and further suggest new perspective of improving text summarization models by sharing our insights on the current learning objective.


Our contributions are as follow:
\begin{itemize}\vspace{-4pt}
\item We evaluate model generated summary and benchmark dataset through human evaluation bases. \footnote{Our evaluation results are available at \url{https://github.com/TBA}}
\item We investigate a benchmark dataset, CNN/DM, and suggest the characteristics of the dataset. We further share our insights on the model-generated summaries. 
\item We present our human evaluation scoring guidelines on 3 criteria; \textit{Creativity}, \textit{Readability} and \textit{Relevance}. 
\vspace{10pt}
\end{itemize}

\section{Related Work}
\subsection{Reinforcement Learning} 
Reinforcement Learning (RL) is a widely used learning technique for text summarization task. 
\citet{paulus2018deep} pointed out that the loss of the supervised model is not closely related to the evaluation metric and therefore, introduced an end-to-end RL model that employs the ROUGE metric \cite{lin-2004-rouge} as a rewarder.

\citet{boehm_emnlp2019_summary_reward} point out the limitations of ROUGE-based rewarders and proposed neural network-based rewarders to predict the similarity between document and summary.
Specifically, the model is trained to predict the similarity score between the document and the summaries of various quality.
The pre-trained language model BERT is used to encode the input sequences so that the semantics of the two inputs are adequately reflected in the model.

\subsection{Supervised Learning}
Supervised Learning is an actively researched area in summarization. 
\citet{see2017get} introduced a sequence-to-seq attentional model that combines \textit{coverage} vector and the copy mechanism.
\citet{gehrmann-etal-2018-bottom} proposed a bottom-up attention model by incorporating the content selection system that selects the important parts of a document.
\citet{liu-lapata-2019-text} presented a document-level encoder using BERT~\cite{devlin2018bert} and showed benefits of using pre-trained language model (LM) as embeddings.
\citet{jeh2020encoderdecoder} developed an approach to stack an additional encoder-decoder network, on top of an attentional encoder-decoder network to alleviate the \textit{exposure bias} issue that comes from teacher forcing~\cite{williams1989learning}.

\paragraph{Pre-trained models}
Recent works on pre-trained language models made significant advances in NLP tasks. 
BERT~\cite{devlin2018bert} is a bidirectional encoder that is pre-trained by predicting the original document with the corrupted document as an input.
GPT~\cite{radford2018improving} and GPT-2~\cite{radford2019language} are auto-regressive LMs. 
BART~\cite{lewis2019bart} is a pre-trained language model that combines bidirectional transformer as an encoder and auto-regressive transformers as a decoder.
Concurrent to our work, ProphetNet~\cite{yan2020prophetnet} and PEGASUS~\cite{zhang2019pegasus} also use pre-trained LMs to solve text summarization task. 
Both model shows stunning performance by using encoder-decoder settings.


\begin{table}[h] 
\resizebox{0.48\textwidth}{!}{
\begin{tabular}{clccc}
\toprule
\multirow{2}{*}{} & \multirow{2}{*}{System} & \multicolumn{3}{c}{\textbf{CNN/DailyMail}}  \\
 & & R-1   & R-2   & R-L  \\
\midrule
\multirow{5}{*}{RL} & \citet{boehm_emnlp2019_summary_reward} & 39.60 & 18.10 & 36.50  \\
& \citet{narayan2018ranking} & 40.00 & 18.20 & 36.60  \\
& \citet{pasunuru2018multi} & 40.43 & 18.00 & 37.10  \\
& \citet{chen2018fast} & 40.88 & 17.80 & 38.54 \\
& \citet{bae2019summary} & 41.90 & 19.08 & 39.64  \\
\midrule
\multirow{4}{*}{SL} & \citet{jeh2020encoderdecoder} & 40.44 & 18.15 & 36.90 \\
& \citet{yan2020prophetnet} & 43.68 & 20.64 & 40.72  \\
& \citet{zhang2019pegasus} & 44.17 & 21.47 & 41.11  \\ 
& BART (Baseline) & 43.98 & 21.07 & 40.82  \\
\bottomrule
\end{tabular}
}
\caption{ROUGE (automatic) evaluations on CNN/DailyMail dataset of reinforcement learning (RL) and supervised learning (SL) models.} \label{tab:auto}
\end{table}

\section{Experiments}

\begin{table*}[h]
\centering{
\begin{tabular}{l|ccc|c}
\toprule
                & Creativity & Readability & Relevance & Total (Avg) \\
\midrule
Reference Summary & 2.70 (56.81\%)      & 3.05 (68.23\%)       & 2.68 (55.96\%)     & 2.81 (60.33\%) \\
BART (Baseline) & 2.80 (60.14\%)      & 3.39 (79.65\%)       & 3.12 (70.71\%)     & 3.10 (70.17\%) \\
\bottomrule
\end{tabular}}
\caption{Human Evaluation score of the Systems. Scores are presented in both 1-4 scale and 0-100\% percent scale (numbers in parenthesis)}\label{tab:humaneval}
\end{table*}

\subsection{Dataset}
We used the non-anonymized CNN/DailyMail (CNN/DM) dataset \citep{hermann2015teaching, see2017get} to evaluate our approach.
CNN/DM dataset is composed of articles and corresponding bullet point summary pairs from the news providers.
Following BART, we applied additional preprocessing steps such as replacing escape characters \footnote{\url{https://github.com/pytorch/fairseq/blob/master/examples/bart/README.cnn.md}}.
The preprocessed CNN/DM dataset includes 287k training pairs, 13k validation pairs, and 11k testing pairs. 
\subsection{Model}

We first introduce the underlying structure of BART in Section~\ref{section:BART}. 

\paragraph{BART}
\label{section:BART}
BART is a denoising autoencoder that uses sequence-to-sequence transformer architecture of \citet{vaswani2017attention}. 
The structure of BART consists of two parts: an encoder and a decoder. 
The encoder part is a bidirectional encoder which corresponds to the structure of BERT \cite{devlin2018bert}, and the decoder part is an auto-regressive decoder following the settings of GPT.

During the pre-training process, BART receives the corrupted document as input and performs the task of predicting the original uncorrupted document. 
In this way, BART can effectively learn contextual representations.

BART can be fine-tuned for various tasks such as token classification, sequence classification and sequence generations. 
When fine-tuned for summarization task, the bidirectional encoder part encodes the original document and the decoder part predicts the reference summary.

We utilize weights of BART large fine-tuned on CNN/DM dataset (\textit{bart-large-cnn}) as our supervised text summarization model \footnote{\url{https://github.com/pytorch/fairseq/tree/master/examples/bart}}.

\subsection{Settings}
Following the setting of original BART paper, we tokenized the input sequences with the byte-pair encoding of RoBERTa \cite{liu2019roberta}.
During the generation process, beam search decoding with a beam size of 4 was used to produce the output summary. Trigram blocking \cite{paulus2018deep}, min-len of 55 tokens, max-len of 140 tokens and length penalty were applied during decoding~\cite{lewis2019bart}.

\section{Evaluations}

\subsection{Automatic Evaluation}

We report our automatic evaluation results on the CNN/DM dataset in Table~\ref{tab:auto}.
For other models, we report the scores in accordance with their papers.
For summarization task, the ROUGE metrics \cite{lin-2004-rouge} are widely used for evaluations, namely F1 scores of ROUGE-1, ROUGE-2 and ROUGE-L~\citep{paulus2018deep, see2017get, lewis2019bart}.
ROUGE score of BART model was better than reference models using RL approach and the most supervised learning models.
Automatic Evaluation results suggest that the BART model is a top-performing model.

\begin{table*}[h]
\resizebox{0.98\textwidth}{!}{
\begin{tabular}{llll}
\toprule
\multicolumn{4}{c}{Relevance} \\ 
\midrule
\multicolumn{1}{c}{1} & \multicolumn{1}{c}{2} & \multicolumn{1}{c}{3} & \multicolumn{1}{c}{4} \\ 
\midrule
\begin{tabular}[c]{@{}l@{}}*No clear overview\\ *Many key features are missed\\ *Lack of information\\ *Inaccurate information\end{tabular} & \begin{tabular}[c]{@{}l@{}}*There is an overview\\ *Some key features are not covered\\ *Inaccurate Information\end{tabular} & \begin{tabular}[c]{@{}l@{}}*A clear overview\\ *Key feature is missed (1 or 2) \\ *Accurate information\end{tabular} & \begin{tabular}[c]{@{}l@{}}*A clear overview\\ *All key features are well \\   illustrated\\ *Accurate information\end{tabular} \\ 
\toprule
\multicolumn{4}{c}{Readability} \\ 
\midrule
\multicolumn{1}{c}{1} & \multicolumn{1}{c}{2} & \multicolumn{1}{c}{3} & \multicolumn{1}{c}{4} \\ 
\midrule
\begin{tabular}[c]{@{}l@{}}*Poorly readable summary \\ *Frequent errors in grammar, \\   punctuation or spelling\\ *Wrong words and informal language\end{tabular} & \begin{tabular}[c]{@{}l@{}}*Rather readable summary\\ *Some errors in grammar, \\   punctuation and spelling\\ *Use less common words\end{tabular} & \begin{tabular}[c]{@{}l@{}}*Easily readable summary\\ *Rare errors in grammar, \\   punctuation or spelling\end{tabular} & \begin{tabular}[c]{@{}l@{}}*Highly readable summary\\ *Sentences are free of errors\\ *No grammar errors\\ *No punctuation errors\\ *No spelling mistakes\end{tabular} \\
\toprule
\multicolumn{4}{c}{Creativity} \\ 
\midrule
\multicolumn{1}{c}{1} & \multicolumn{1}{c}{2} & \multicolumn{1}{c}{3} & \multicolumn{1}{c}{4} \\ 
\midrule
\begin{tabular}[c]{@{}l@{}}*Completely same as the "original"\\ *Summary is from the beginning part \\   of the "original"\end{tabular} & \begin{tabular}[c]{@{}l@{}}*No attempt to create sentences\\ *Copied most sentences from \\   the "original"\\ *Most sentences are from the \\   front part of the "original"\\ *Poor understanding of collocations\end{tabular} & \begin{tabular}[c]{@{}l@{}}*Some sentences are generated \\   but have inaccurate meaning\\ *Tries to use complex sentences \\   with limited success\\ *Most sentences are from the \\   "original"\\ *Some sentences are from the \\   front part of the "original"\end{tabular} & \begin{tabular}[c]{@{}l@{}}*Creatively used a range of \\   vocabulary to generate summary\\   (Relatively few sentences are \\   from the "original")\\ *Has precise meaning\\ *Understand collocations\\ *Use referencing or linking \\ words (ex, this, it, and, however etc.)\end{tabular} \\
\bottomrule
\end{tabular}}
\caption{Human evaluation criteria. Higher score indicates better quality of summary.}
\label{tab:cri}
\end{table*}

\subsection{Human Evaluation}

\subsubsection{Evaluation Criteria}

In order to assess model’s proficiency in summarization, we follow the evaluation criteria of International English Language Testing System (IELTS)\footnote{\url{https://www.ielts.org}} as it is one of the major English test for non-native speakers across the world. 
IELTS writing is about summarizing information in a graph or table and writing a letter in response to a problem. 
Although it has different nature to the summarization task, we believe the fundamental factors should be the same because the model also needs to comprehend the given information, grasp important matters, and write with its own words. 

We modified evaluation criteria to Relevance, Readability and Creativity. 
Both Relevance and Readability are referred from IELTS’ criteria but Creativity is added specifically for sentence summarization task.
Creativity is a meaningful factor because a good summary should not be copied from the original text, but rather translated into the model’s own words to represent the context. 

The following questions are asked to adopt score guidelines for our experiment: 
\begin{itemize}
    \item \textit{Creativity}- Is the summary written with its own words and sentence structures?
    \item \textit{Readability}- Does the summary avoid grammar errors and informal language?
    \item \textit{Relevance}- Does the summary contain both important and accurate information about the original document?
\end{itemize}
For more descriptions about score guidelines, please refer to Table~\ref{tab:cri}.

\subsubsection{Evaluation Setup}\label{section:evaluationsetup}

We use Amazon Mechanical Turk to evaluate the machine-generated summaries. 
For qualifications, we required all workers to possess a bachelor’s degree in the United States. 
Then, we organized a team into ten workers, and each team is requested to answer five questions, where one question includes the original document, the reference summary, and BART model generated summaries.
The reference summary and model generated summary are presented in random order. 
Overall, ten teams are participated for evaluation, meaning 100 people (1 team x 10 people) and 50 examples (5 examples x 10 teams) in total.

Workers are asked to measure the level of summarization quality from 1 to 4 in terms of Relevance, Readability, and Creativity. 
For these three criteria, the human examiner will judge whether the summary contains key features, avoids grammar errors, and uses its own words and sentence structures. 
Please consult Appendix for further details about score requirements.

Turk workers in our experiment had Human Intelligence Task (HIT) approval rate of 98\% and completed over 10,000 HITs. 
For five questions, they spent about 1437.69 seconds(24min) on average, and the standard deviation is 782.4 seconds(13min). 
These statistics suggest that the time it took for workers to complete the questions varied greatly. 
It is clear that some workers did not respond faithfully, and therefore we excluded workers whose spent time is in the lower 5\%, which is 389 seconds. 
Upon inspection, we realize that these inattentive workers are not biased to a certain team. 
Instead, there are about 0 to 2 workers in each team, whose spent time is less than 389 seconds for five questions. 
In this manner, we tried to not only ensure fairness but also assure the quality of human evaluation.

\subsubsection{Results}
Our human evaluation results are reported in Table~\ref{tab:humaneval}.
The table shows averaged score across all 470 responses (5 examples X 94 people). 
We convert the human evaluation scores from 1-4 scale to 0-100 scale and reported total score, which is the average of three scores.
We compare the scores of BART model and the reference summary.
BART was better than reference summaries in terms of Readability (p-value $< 10^{-10}$) and Relevance (p-value $< 10^{-10}$), but the gain in Creativity was marginal.
We calculated p-values using the one-tailed t-test (statistical significance of 0.01).

\begin{table*}[h]  
\resizebox{\textwidth}{!}{
\begin{tabular}{lc}
\hline
 & CNN/DailyMail dataset~ ~ ~ ~ ~~ ~ ~ ~ ~ ~~ ~ ~ ~ ~  \\ \hline
\multicolumn{1}{l|}{Document} & \begin{tabular}[c]{@{}l@{}}
Brazilian supermodel Alessandra Ambrosio goes back to her roots in an edgy new campaign shot \\
in her home country. The 34-year-old Victoria's Secret Angel shows off her Latino style and \\
golden tan as she poses in a new campaign for online fashion retailer Dafiti, shot in São Paulo. \\
Dafiti, Latin America’s largest online fashion retailer, has launched its own fashion collection,\\
the Dafiti Collection, and signed Alessandra because they believe she embodies the style of the \\
brand. . Scroll down for video . Alessandra Ambrosio, who found fame as a Victoria's Secret \\
Angel, has been snapped up to front a campaign for online fashion retailer Dafiti, which was shot \\
in São Paulo. The mother-of-two, who is No. 8 on the Forbes list of top-earning models, stars in \\
the advertising ~~\ldots ~~
 \end{tabular}  \\ \hline
\multicolumn{1}{l|}{Reference Summary} & \begin{tabular}[c]{@{}l@{}} 

Mother-of-two, 34, snapped up to front Dafiti's AW15 campaign. Latin American e-tailer believe \\
she embodies the style of the brand. Recently named No. 8 on Forbes list of top-earning models.
\end{tabular} \\ \hline
\end{tabular}}
\caption{An example of CNN/DailyMail dataset.} \label{tab:exmcnn}
\end{table*}

\begin{figure}[h]
\centering
\fbox{
\includegraphics[width=\columnwidth]{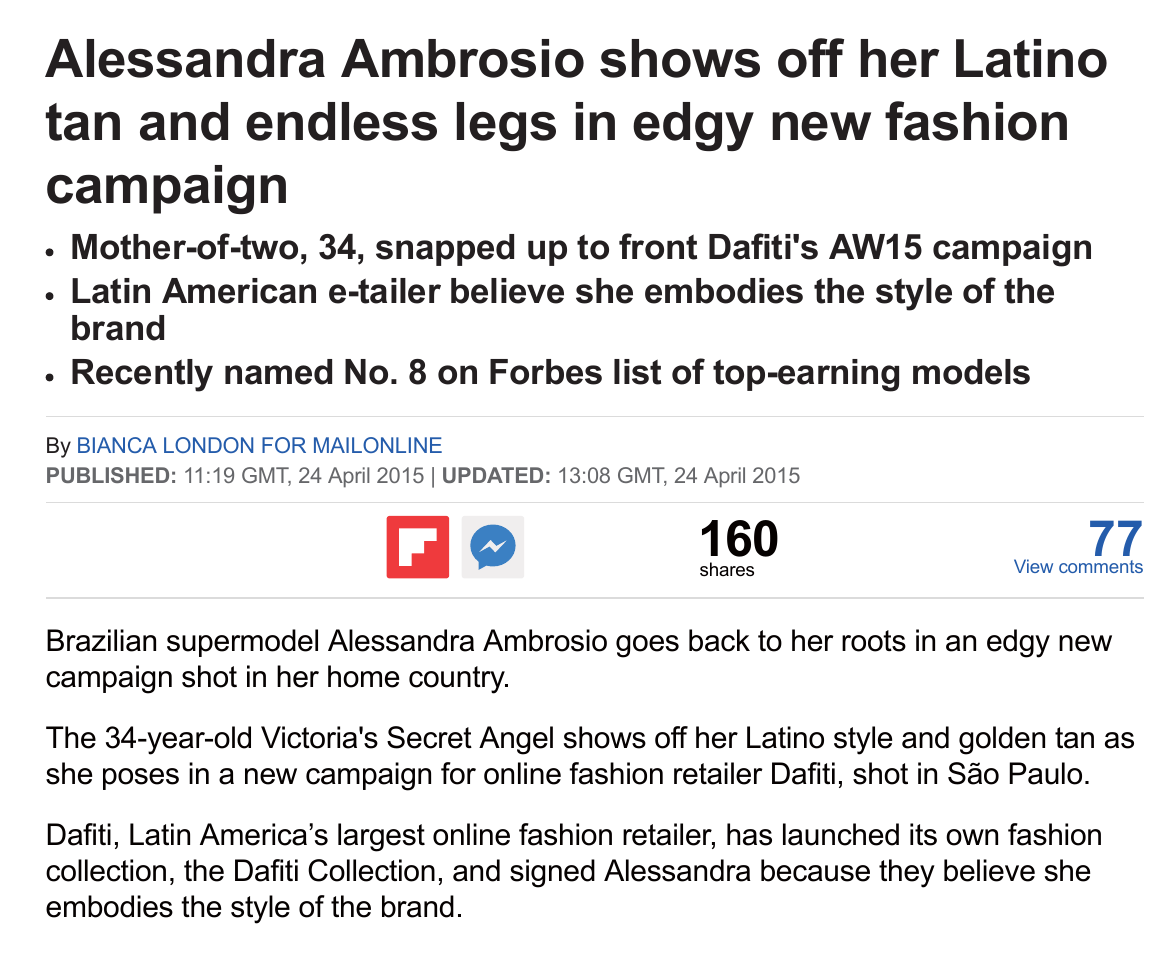}
}
\caption{An example of CNN/DailyMail dataset.} \label{fig:cnndmexam}
\end{figure}

\section{Discussion} \label{section:discus}

\subsection{Results Analysis}
It is an interesting observation that a model produced a more favorable summaries than the reference summaries.
In this section, we take a deeper look in to this phenomenon, by focusing on two following aspects: 1) The reference summary is not always an ideal summary of the document and 2) A large-scale language model has a strong ability in text generation.

As a nature of the CNN/DM dataset, some reference summaries are of poor quality and hence, these summaries received low scores compared to those generated by models. Table~\ref{tab:exmcnn} and Figure~\ref{fig:cnndmexam} show an example of a document-summary pair. This example summary is missing crucial information to comprehend the document: the subject of the article \textit{Alessandra Ambrosio} and who she is. For this example, reference summaries are made up of bullet points that are appeared as part of the headlines~(Figure~\ref{fig:cnndmexam}). Headlines are often designed to intrigue readers’ attention, and therefore they usually do not contain enough facts to understand the articles. It is a fatal flaw if the summary is not complete and consequently, such a summary will receive a low score for Readability. We believe that this is one of the reasons why some reference summaries received lower scores than generated summaries.

In general, we assume that the test set and training set are of the same quality because the test set intrinsically has a very similar distribution to the training set. If the test set is a low-quality data, then the training set, which is used for training, would be low quality as well. And this low quality will be reflected eventually when we train a model. Hence, it is rare for the model to produce results that are better than the dataset.
However, our model has shown contradicting results: model's generated summaries are evaluated as better than the reference summaries. 
We believe that this is due to the transfer learning of large-scale language models. Language models are pre-trained on various corpus and we exploited this advantage. 
We assume that the performance difference comes from the language model. 
The idea of the language model is to learn the general understanding of a language during the pre-training process and the specific task during the fine-tuning process. 
BART is trained to extract salient information by using the encoder and to make a complete sentence by using the decoder. 
Generating a complete sentence is closely related to providing a readable sentence and this can be achieved by extracting salient information, which can enhance relevance. 
For these reasons, as shown in the Table~\ref{tab:humaneval}, BART shows outstanding performance in both Readability and Relevance. The successful use of pre-trained knowledge is the main reason why our model and BART received favorable scores. In addition, by considering that the human evaluation score is subjective, the Readability score of almost 80 is regarded as a high score.


\subsection{Learning Strategies and Evaluation Criteria}
In this section, we first share our insights on the model-generated summaries. We aim to open a discussion for a better learning strategy and also like to emphasize a need for a better evaluation criteria for text generation tasks.

\begin{figure}[h]
\centering
\includegraphics[width=\columnwidth]{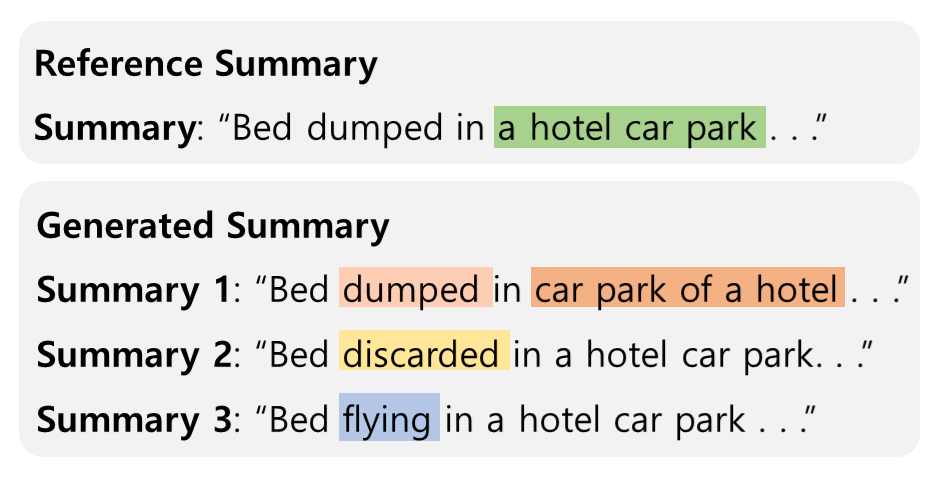}
\caption{Example sentences from summaries} \label{fig:run_ex}
\end{figure}

For sequence generation task, auto-regressive decoder models are trained to predict the next token given previous tokens. 
They mostly use maximum-likelihood method as a training objective to minimize loss when the input and output sequence are identical.

However, we hypothesize that this maximum-likelihood method, which minimizes cross-entropy (CE) loss, is not appropriate for summarization. 
More specifically, the conventional CE loss have the risk of mishandling potentially valid summaries, which are semantically analogous to reference summaries, by considering them as wrong predictions.
Since existing supervised learning models are trained to reproduce the reference summary exactly, producing analogous summaries will count toward wrong prediction and this phenomenon will eventually harm the training process. 
This results from summarization being an open-ended task and maximum-likelihood method being too strict to account for multiple valid answers.

For instance, consider the example sentences in Figure~\ref{fig:run_ex}.
Although the generated summary 1 is semantically parallel to the reference summary, the model receives penalties for the tokens “car park of a hotel” from the generated summary 1. 
Similarly, even though the generated summary 2 is closely related to the reference summary, the model mishandles this sentence as it does not include the word “dumped”. 

\paragraph{Learning Strategy} From this insight, we suggest a better learning strategy can further unlock the potential of text generation using non-RL based supervised learning.
As an example, by maximising semantic similarity as training objective, instead of cross-entropy loss, a model can obtain flexibility and handle multiple valid answers, which are otherwise penalized to act as a training noise.

\paragraph{Evaluation Criteria} The example sentences in Figure~\ref{fig:run_ex} also put emphasis on better evaluation criteria. 
Current n-gram based evaluation criteria, such as ROUGE, is a effective and practical method, yet it may mislead the performance of text generation model by considering a valid answer (or n-gram) as a wrong one.
Hence, we remain a new evaluation criteria which mainly focuses on semantic similarity, as a future work.

\section*{Author's note}
The initial version of the manuscript includes a model design (semsim), experimental results on our model, BART model, and the reference summaries (both automatic evaluation metric and human evaluation metric), and discussions on the results. After we archived the manuscript, we found that our model has flaws in its implementation and design. 
This final version of the manuscript is from the rest of the initial paper; we included our findings on the benchmark dataset, BART generated results and human evaluations, and we excluded our model semsim.


\bibliographystyle{acl_natbib}
\bibliography{emnlp2020}

\begin{thebibliography}{29}
\expandafter\ifx\csname natexlab\endcsname\relax\def\natexlab#1{#1}\fi

\bibitem[{Bae et~al.(2019)Bae, Kim, Kim, and Lee}]{bae2019summary}
Sanghwan Bae, Taeuk Kim, Jihoon Kim, and Sang-goo Lee. 2019.
\newblock Summary level training of sentence rewriting for abstractive
  summarization.
\newblock In \emph{Proceedings of the 2nd Workshop on New Frontiers in
  Summarization}, pages 10--20.

\bibitem[{B{\" o}hm et~al.(2019)B{\" o}hm, Gao, Meyer, Shapira, Dagan, and
  Gurevych}]{boehm_emnlp2019_summary_reward}
Florian B{\" o}hm, Yang Gao, Christian~M. Meyer, Ori Shapira, Ido Dagan, and
  Iryna Gurevych. 2019.
\newblock Better rewards yield better summaries: Learning to summarise without
  references.
\newblock In \emph{Proceedings of the 2019 Conference on Conference on
  Empirical Methods in Natural Language Processing {(EMNLP)}}, Hong Kong,
  China.

\bibitem[{Chen et~al.(2018)Chen, Dai, Tao, Zhang, Gan, Shen, Zhang, Wang,
  Zhang, and Carin}]{chen2018adversarial}
Liqun Chen, Shuyang Dai, Chenyang Tao, Haichao Zhang, Zhe Gan, Dinghan Shen,
  Yizhe Zhang, Guoyin Wang, Ruiyi Zhang, and Lawrence Carin. 2018.
\newblock Adversarial text generation via feature-mover's distance.
\newblock In \emph{Advances in Neural Information Processing Systems}, pages
  4666--4677.

\bibitem[{Chen and Bansal(2018)}]{chen2018fast}
Yen-Chun Chen and Mohit Bansal. 2018.
\newblock Fast abstractive summarization with reinforce-selected sentence
  rewriting.
\newblock In \emph{Proceedings of the 56th Annual Meeting of the Association
  for Computational Linguistics (Volume 1: Long Papers)}, pages 675--686.

\bibitem[{Devlin et~al.(2018)Devlin, Chang, Lee, and
  Toutanova}]{devlin2018bert}
Jacob Devlin, Ming-Wei Chang, Kenton Lee, and Kristina Toutanova. 2018.
\newblock Bert: Pre-training of deep bidirectional transformers for language
  understanding.
\newblock \emph{arXiv preprint arXiv:1810.04805}.

\bibitem[{Gehrmann et~al.(2021)Gehrmann, Adewumi, Aggarwal, Ammanamanchi,
  Anuoluwapo, Bosselut, Chandu, Clinciu, Das, Dhole et~al.}]{gehrmann2021gem}
Sebastian Gehrmann, Tosin Adewumi, Karmanya Aggarwal, Pawan~Sasanka
  Ammanamanchi, Aremu Anuoluwapo, Antoine Bosselut, Khyathi~Raghavi Chandu,
  Miruna Clinciu, Dipanjan Das, Kaustubh~D Dhole, et~al. 2021.
\newblock The gem benchmark: Natural language generation, its evaluation and
  metrics.
\newblock \emph{arXiv preprint arXiv:2102.01672}.

\bibitem[{Gehrmann et~al.(2018)Gehrmann, Deng, and
  Rush}]{gehrmann-etal-2018-bottom}
Sebastian Gehrmann, Yuntian Deng, and Alexander Rush. 2018.
\newblock \href {https://doi.org/10.18653/v1/D18-1443} {Bottom-up abstractive
  summarization}.
\newblock In \emph{Proceedings of the 2018 Conference on Empirical Methods in
  Natural Language Processing}, pages 4098--4109, Brussels, Belgium.
  Association for Computational Linguistics.

\bibitem[{Hermann et~al.(2015)Hermann, Kocisky, Grefenstette, Espeholt, Kay,
  Suleyman, and Blunsom}]{hermann2015teaching}
Karl~Moritz Hermann, Tomas Kocisky, Edward Grefenstette, Lasse Espeholt, Will
  Kay, Mustafa Suleyman, and Phil Blunsom. 2015.
\newblock Teaching machines to read and comprehend.
\newblock In \emph{Advances in neural information processing systems}, pages
  1693--1701.

\bibitem[{Jeh(2020)}]{jeh2020encoderdecoder}
Glen Jeh. 2020.
\newblock \href {https://openreview.net/forum?id=SylkzaEYPS} {Encoder-decoder
  network as loss function for summarization}.

\bibitem[{Lewis et~al.(2019)Lewis, Liu, Goyal, Ghazvininejad, Mohamed, Levy,
  Stoyanov, and Zettlemoyer}]{lewis2019bart}
Mike Lewis, Yinhan Liu, Naman Goyal, Marjan Ghazvininejad, Abdelrahman Mohamed,
  Omer Levy, Ves Stoyanov, and Luke Zettlemoyer. 2019.
\newblock Bart: Denoising sequence-to-sequence pre-training for natural
  language generation, translation, and comprehension.
\newblock \emph{arXiv preprint arXiv:1910.13461}.

\bibitem[{Lin(2004)}]{lin-2004-rouge}
Chin-Yew Lin. 2004.
\newblock \href {https://www.aclweb.org/anthology/W04-1013} {{ROUGE}: A package
  for automatic evaluation of summaries}.
\newblock In \emph{Text Summarization Branches Out}, pages 74--81, Barcelona,
  Spain. Association for Computational Linguistics.

\bibitem[{Liu and Lapata(2019)}]{liu-lapata-2019-text}
Yang Liu and Mirella Lapata. 2019.
\newblock \href {https://doi.org/10.18653/v1/D19-1387} {Text summarization with
  pretrained encoders}.
\newblock In \emph{Proceedings of the 2019 Conference on Empirical Methods in
  Natural Language Processing and the 9th International Joint Conference on
  Natural Language Processing (EMNLP-IJCNLP)}, pages 3730--3740, Hong Kong,
  China. Association for Computational Linguistics.

\bibitem[{Liu et~al.(2019)Liu, Ott, Goyal, Du, Joshi, Chen, Levy, Lewis,
  Zettlemoyer, and Stoyanov}]{liu2019roberta}
Yinhan Liu, Myle Ott, Naman Goyal, Jingfei Du, Mandar Joshi, Danqi Chen, Omer
  Levy, Mike Lewis, Luke Zettlemoyer, and Veselin Stoyanov. 2019.
\newblock Roberta: A robustly optimized bert pretraining approach.
\newblock \emph{arXiv preprint arXiv:1907.11692}.

\bibitem[{Maynez et~al.(2020)Maynez, Narayan, Bohnet, and
  McDonald}]{maynez2020faithfulness}
Joshua Maynez, Shashi Narayan, Bernd Bohnet, and Ryan McDonald. 2020.
\newblock On faithfulness and factuality in abstractive summarization.
\newblock \emph{arXiv preprint arXiv:2005.00661}.

\bibitem[{Narayan et~al.(2018)Narayan, Cohen, and Lapata}]{narayan2018ranking}
Shashi Narayan, Shay~B Cohen, and Mirella Lapata. 2018.
\newblock Ranking sentences for extractive summarization with reinforcement
  learning.
\newblock In \emph{Proceedings of the 2018 Conference of the North American
  Chapter of the Association for Computational Linguistics: Human Language
  Technologies, Volume 1 (Long Papers)}, pages 1747--1759.

\bibitem[{Pasunuru and Bansal(2018)}]{pasunuru2018multi}
Ramakanth Pasunuru and Mohit Bansal. 2018.
\newblock Multi-reward reinforced summarization with saliency and entailment.
\newblock In \emph{Proceedings of the 2018 Conference of the North American
  Chapter of the Association for Computational Linguistics: Human Language
  Technologies, Volume 2 (Short Papers)}, pages 646--653.

\bibitem[{Paulus et~al.(2018)Paulus, Xiong, and Socher}]{paulus2018deep}
Romain Paulus, Caiming Xiong, and Richard Socher. 2018.
\newblock A deep reinforced model for abstractive summarization.

\bibitem[{Peters et~al.(2018)Peters, Neumann, Iyyer, Gardner, Clark, Lee, and
  Zettlemoyer}]{Peters:2018}
Matthew~E. Peters, Mark Neumann, Mohit Iyyer, Matt Gardner, Christopher Clark,
  Kenton Lee, and Luke Zettlemoyer. 2018.
\newblock Deep contextualized word representations.
\newblock In \emph{Proc. of NAACL}.

\bibitem[{Radford et~al.(2018)Radford, Narasimhan, Salimans, and
  Sutskever}]{radford2018improving}
Alec Radford, Karthik Narasimhan, Tim Salimans, and Ilya Sutskever. 2018.
\newblock Improving language understanding by generative pre-training.
\newblock \emph{URL https://s3-us-west-2. amazonaws.
  com/openai-assets/researchcovers/languageunsupervised/language understanding
  paper. pdf}.

\bibitem[{Radford et~al.(2019)Radford, Wu, Child, Luan, Amodei, and
  Sutskever}]{radford2019language}
Alec Radford, Jeffrey Wu, Rewon Child, David Luan, Dario Amodei, and Ilya
  Sutskever. 2019.
\newblock Language models are unsupervised multitask learners.
\newblock \emph{OpenAI Blog}, 1(8):9.

\bibitem[{Schluter(2017)}]{schluter2017limits}
Natalie Schluter. 2017.
\newblock The limits of automatic summarisation according to rouge.
\newblock In \emph{Proceedings of the 15th Conference of the European Chapter
  of the Association for Computational Linguistics: Volume 2, Short Papers},
  pages 41--45.

\bibitem[{See et~al.(2017)See, Liu, and Manning}]{see2017get}
Abigail See, Peter~J Liu, and Christopher~D Manning. 2017.
\newblock Get to the point: Summarization with pointer-generator networks.
\newblock \emph{arXiv preprint arXiv:1704.04368}.

\bibitem[{Vaswani et~al.(2017)Vaswani, Shazeer, Parmar, Uszkoreit, Jones,
  Gomez, Kaiser, and Polosukhin}]{vaswani2017attention}
Ashish Vaswani, Noam Shazeer, Niki Parmar, Jakob Uszkoreit, Llion Jones,
  Aidan~N Gomez, {\L}ukasz Kaiser, and Illia Polosukhin. 2017.
\newblock Attention is all you need.
\newblock In \emph{Advances in neural information processing systems}, pages
  5998--6008.

\bibitem[{Williams and Zipser(1989)}]{williams1989learning}
Ronald~J Williams and David Zipser. 1989.
\newblock A learning algorithm for continually running fully recurrent neural
  networks.
\newblock \emph{Neural computation}, 1(2):270--280.

\bibitem[{Yan et~al.(2020)Yan, Qi, Gong, Liu, Duan, Chen, Zhang, and
  Zhou}]{yan2020prophetnet}
Yu~Yan, Weizhen Qi, Yeyun Gong, Dayiheng Liu, Nan Duan, Jiusheng Chen, Ruofei
  Zhang, and Ming Zhou. 2020.
\newblock Prophetnet: Predicting future n-gram for sequence-to-sequence
  pre-training.
\newblock \emph{arXiv preprint arXiv:2001.04063}.

\bibitem[{You et~al.(2019)You, Jia, Liu, and Yang}]{you-etal-2019-improving}
Yongjian You, Weijia Jia, Tianyi Liu, and Wenmian Yang. 2019.
\newblock \href {https://doi.org/10.18653/v1/P19-1205} {Improving abstractive
  document summarization with salient information modeling}.
\newblock In \emph{Proceedings of the 57th Annual Meeting of the Association
  for Computational Linguistics}, pages 2132--2141, Florence, Italy.
  Association for Computational Linguistics.

\bibitem[{Zhang et~al.(2019{\natexlab{a}})Zhang, Cai, Xu, and
  Wang}]{zhang-etal-2019-pretraining}
Haoyu Zhang, Jingjing Cai, Jianjun Xu, and Ji~Wang. 2019{\natexlab{a}}.
\newblock \href {https://doi.org/10.18653/v1/K19-1074} {Pretraining-based
  natural language generation for text summarization}.
\newblock In \emph{Proceedings of the 23rd Conference on Computational Natural
  Language Learning (CoNLL)}, pages 789--797, Hong Kong, China. Association for
  Computational Linguistics.

\bibitem[{Zhang et~al.(2019{\natexlab{b}})Zhang, Zhao, Saleh, and
  Liu}]{zhang2019pegasus}
Jingqing Zhang, Yao Zhao, Mohammad Saleh, and Peter~J Liu. 2019{\natexlab{b}}.
\newblock Pegasus: Pre-training with extracted gap-sentences for abstractive
  summarization.
\newblock \emph{arXiv preprint arXiv:1912.08777}.

\bibitem[{Zhang et~al.(2018)Zhang, Lapata, Wei, and
  Zhou}]{zhang-etal-2018-neural}
Xingxing Zhang, Mirella Lapata, Furu Wei, and Ming Zhou. 2018.
\newblock \href {https://doi.org/10.18653/v1/D18-1088} {Neural latent
  extractive document summarization}.
\newblock In \emph{Proceedings of the 2018 Conference on Empirical Methods in
  Natural Language Processing}, pages 779--784, Brussels, Belgium. Association
  for Computational Linguistics.

\end{thebibliography}


\end{document}